\documentclass[10pt,twocolumn,letterpaper]{article}

\usepackage{iccv}
\usepackage{times}
\usepackage{epsfig}
\usepackage{graphicx}
\usepackage{amsmath}
\usepackage{amssymb}
\usepackage{pifont}
\usepackage{subfigure}
\usepackage{algorithm}
\usepackage{algorithmic}
\graphicspath{{./figures/}}

\newcommand{\modelname}{ADA-CM }

\usepackage[breaklinks=true,colorlinks,bookmarks=false]{hyperref}

\iccvfinalcopy 

\def\iccvPaperID{****} 

\ificcvfinal\fi

\begin{document}

\title{Efficient Adaptive Human-Object Interaction Detection with\\ Concept-guided Memory}

\author{Ting Lei$^1$\quad Fabian Caba$^2$\quad Qingchao Chen$^3$\quad Hailin Jin$^2$\quad Yuxin Peng$^1$\quad Yang Liu$^{1}$\thanks{Corresponding author} \\
$^1$Wangxuan Institute of Computer Technology, Peking University\\$^2$Adobe Research\quad $^3$National Institute of Health Data Science, Peking University\\
{\tt\small \{ting\_lei, qingchao.chen, pengyuxin, yangliu\}@pku.edu.cn}\\
{\tt\small \{caba, hljin\}@adobe.com}
}

\maketitle
\ificcvfinal\thispagestyle{empty}\fi

\begin{abstract}
   Human Object Interaction (HOI) detection aims to localize and infer the relationships between a human and an object. Arguably, training supervised models for this task from scratch presents challenges due to the performance drop over rare classes and the high computational cost and time required to handle long-tailed distributions of HOIs in complex HOI scenes in realistic settings. This observation motivates us to design an HOI detector that can be trained even with long-tailed labeled data and can leverage existing knowledge from pre-trained models. Inspired by the powerful generalization ability of the large Vision-Language Models (VLM) on classification and retrieval tasks, we propose an efficient Adaptive HOI Detector with Concept-guided Memory (ADA-CM). ADA-CM has two operating modes. The first mode makes it tunable without learning new parameters in a training-free paradigm. Its second mode incorporates an instance-aware adapter mechanism that can further efficiently boost performance if updating a lightweight set of parameters can be afforded. Our proposed method achieves competitive results with state-of-the-art on the HICO-DET and V-COCO datasets with much less training time. Code can be found at \url{https://github.com/ltttpku/ADA-CM}.
\end{abstract}

\begin{figure}[!ht]
  \centering

  \subfigure[\textbf{Performance comparisons on rare and non-rare HOI classes.} 
  \modelname performs well on rare and non-rare classes in both the training-free and fine-tuning settings.
  ]{
  \includegraphics[width=0.85\linewidth]{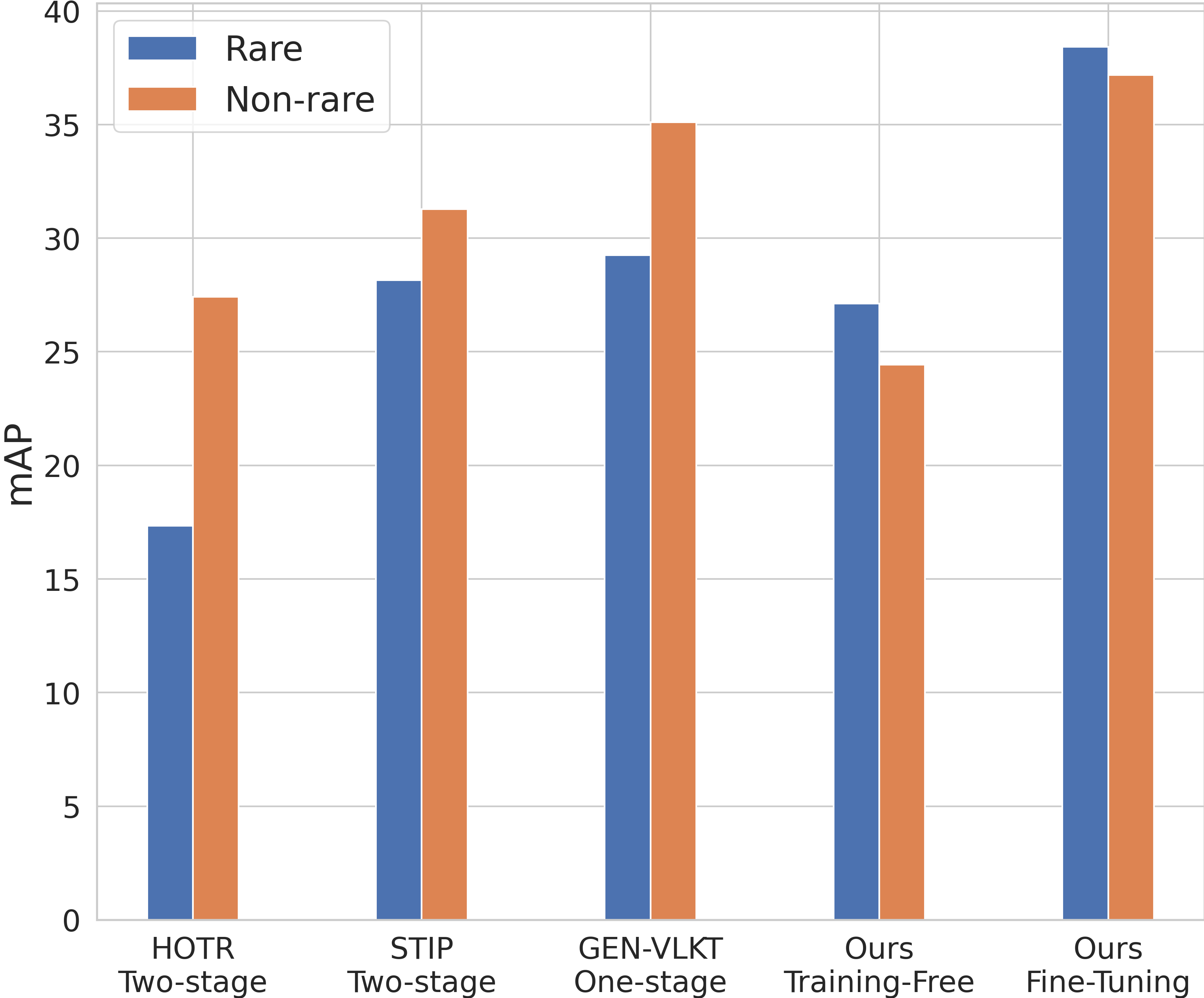}
   \label{fig:rare_nonrare}
  }
  

  \subfigure[\textbf{Performance \textit{vs} Efficiency analysis on HICO-DET dataset.} 
  The size of the blobs is proportional to the models' training gpu time, spanning from 14.85 to 163.7 hours.  
  ]{
  \includegraphics[width=0.85\linewidth]{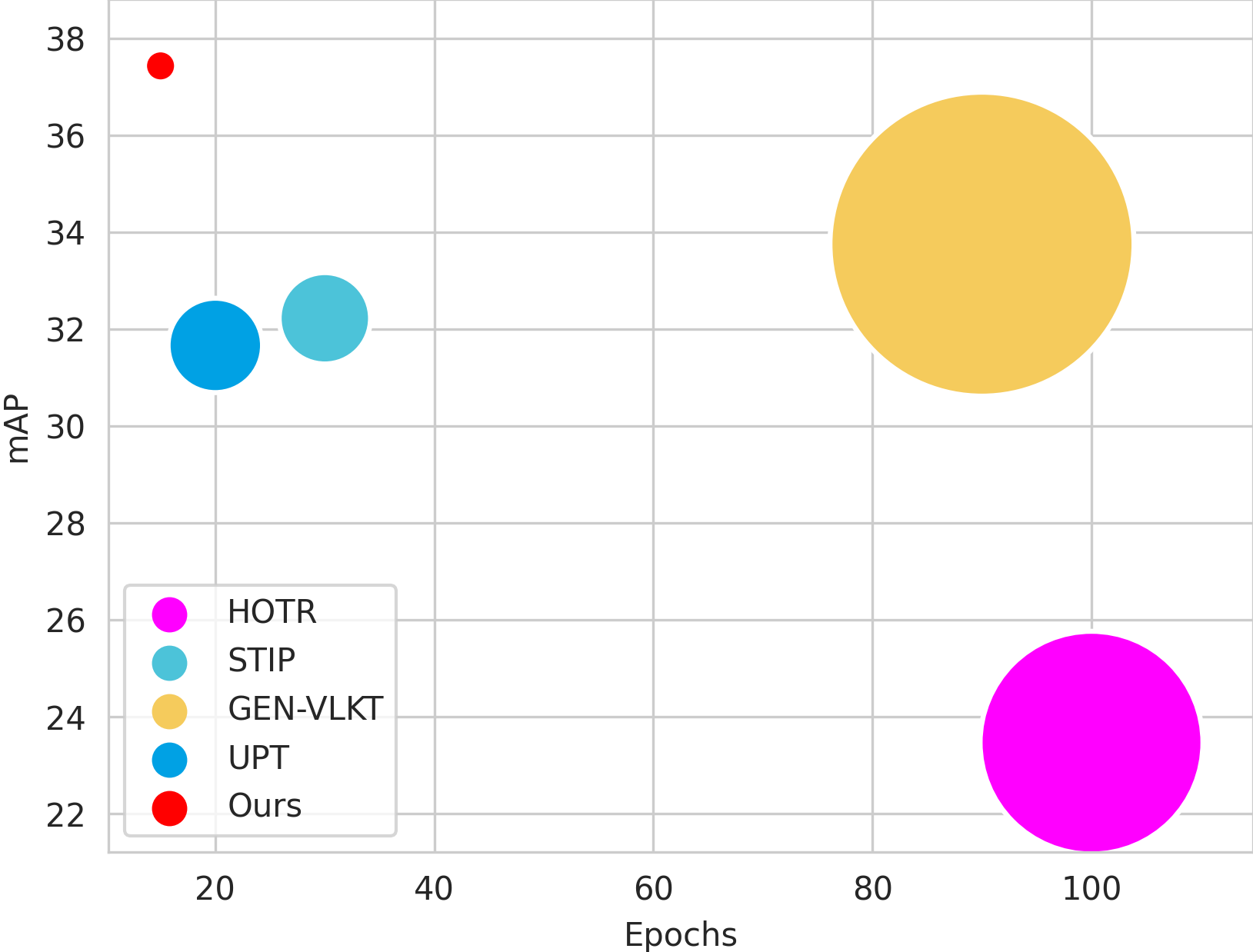}
  \label{fig:scatter}
  }
  
\caption{\textbf{Comparisons on HICO-DET dataset.} Our model, \modelname, achieves an mAP of 37.47\%, performing well on rare HOI classes as shown in~\ref{fig:rare_nonrare}. Compared to other HOI detectors, our method offers tremendous benefits in terms of accuracy and efficiency at training time. }
\label{fig:efficiency}
\end{figure}

\section{Introduction}
\label{sec:intro}

Human-object interaction (HOI) detection is essential for comprehending human-centric scenes at a high level.
Given an image, HOI detection aims to localize human and object pairs and recognize their interactions, \textit{i.e.} a set of $<$human, object, action$>$ triplets. 
Recently, vision Transformers~\cite{vaswani2017attention}, especially the DEtection TRansformer (DETR)~\cite{carion2020end}, have started to revolutionize the HOI detection task. 
Two-stage methods use an off-the-shelf detector, \eg DETR, to localize humans and objects concurrently, followed by predicting interaction classes using the localized region features. One-stage methods usually leverage the pre-trained or fine-tuned weights and architecture of DETR to predict HOI triplets from a global image context in an end-to-end manner. 

Despite the progress, most previous methods still face two major challenges in solving the HOI detection task. 
\textbf{First}, annotating HOI pairs requires considerable human effort. 
Therefore, the model may experience data scarcity when encountering a new domain.
Even with the availability of ample data, the combinatorial nature of HOIs exacerbates challenging scenarios such as recognizing HOI classes from long-tailed distributions. While people can efficiently learn to recognize seen and even unseen HOI from limited samples, most previous methods ~\cite{kim2021hotr,tamura2021qpic, liao2022gen} suffer a significant performance drop on rare classes as shown in Figure~\ref{fig:rare_nonrare}.
\textbf{Second}, training or fine-tuning an HOI detector can contribute to high computational cost and time as shown in Figure~\ref{fig:scatter}\footnote{We exclude the training time for pre-training or fine-tuning the object detector since all methods follow the same protocal to pre-train or fine-tune the object detector.}. Training two-stage methods involves exhaustively combining instance-level features to predict pairwise relationships. In constast, training one-stage HOI detectors that adopt the architecture of DETR model can be challenging due to the heavy reliance on transformers~\cite{liu2020understanding}.

According to the challenges mentioned above, our goal is to \textit{build an efficient adaptive HOI detector resilient to imbalanced data, which can not only adapt to a target dataset without training but also quickly converge when fine-tuning.}

To deal with the problem of lacking labeled data in a target HOI visual domain, we propose a training-free approach with a concept-guided memory module that provides a balanced memory mechanism for all HOI classes.
The concept-guided memory module leverages not only the \textit{domain-specific visual knowledge}, but also the \textit{domain-agnostic semantic knowledge} for HOI detection task. 
Specifically, we extract features from the identified regions of interest to create domain-specific visual knowledge.
Inspired by the impressive zero-shot capability of large visual-language models~\cite{radford2021learning}, we extract the semantic embeddings of HOI categories and treat the language prior as the domain-agnostic semantic knowledge. This enables the model to leverage linguistic commonsense to capture possible co-occurrences and affordance relations between objects and interactions. 
To store and retrieve the knowledge, we propose to construct a key-value concept-guided memory module, which can help to mitigate the problem of data scarcity by providing additional prior information and guidance to the model.
As shown in Figure~\ref{fig:rare_nonrare}, our approach can work well in a \textit{training-free} manner and can effectively detect rare HOI classes.

Moreover, to quickly adapt to new domains, we propose to unfreeze the properly initialized concept-guided cache memory and \textit{inject lightweight residual instance-aware adapters} at spatial sensitive feature maps during training. 
Specifically, unfreezing the cache memory enables the model to select which knowledge is highlighted or suppressed dynamically for the HOI task. Since many valuable cues for the HOI detection task may appear from the early spatial-aware and fine-grained feature maps, we propose to early inject prior knowledge into low-level feature maps to capture the local geometric spatial structures required for pair-wise relationship detection. 
During fine-tuning, while the instance-aware adapters are tailored to facilitate HOI understanding given instance-level prior knowledge, the explicit concept-guided memory mechanism can help alleviate forgetfulness of rare HOI classes as shown in Figure~\ref{fig:rare_nonrare}. 
Furthermore, Figure~\ref{fig:scatter} demonstrates that our network can achieve the best performance by training for a few epochs with fast convergence speed.

\noindent\textbf{Contributions.} Our key idea is to design a HOI detector that leverages knowledge from pre-trained vision-language models and can be adapted to new domains via training-free or fine-tuning. Our work brings three contributions:\\
\textbf{(1)} To the best of our knowledge, we are the first to propose a training-free human-object interaction detection approach, by constructing a balanced concept-guided memory that leverages domain-specific visual knowledge and domain-agnostic semantic knowledge.\\
\textbf{(2)} We demonstrate that unfreezing the properly initialized cache memory and injecting lightweight residual instance-aware adapters at spatial sensitive feature maps during training further boost the performance.\\
\textbf{(3)} Our approach achieves competitive results on VCOCO \cite{gupta2015visual} and state-of-the-art on HICO-DET \cite{chao2018learning} dataset by training for a few epochs with fast convergence speed.

\begin{figure*}
\begin{center}
\includegraphics[width=0.95\linewidth]{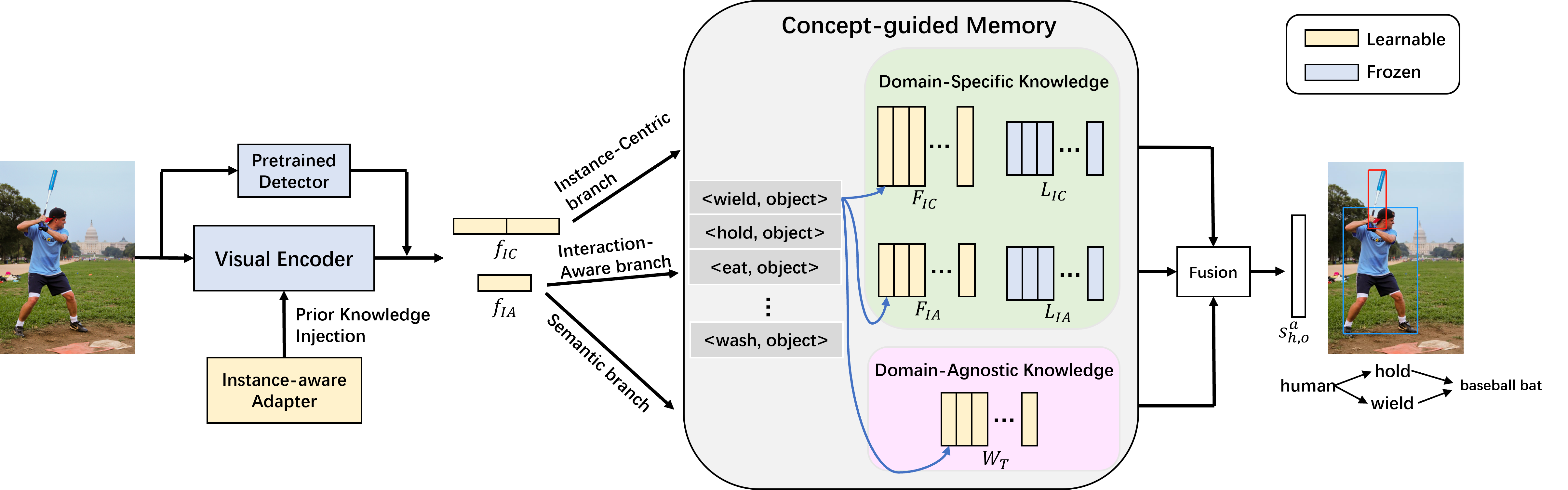}
\end{center}
    \caption{\textbf{The overall framework of our \modelname.} The proposed method supports two settings: \textit{training-free} and \textit{fine-tuning}. For both, human, object and union features are obtained from training samples through CLIP and converted to instance-centric features $f_{IC}$ and interaction-aware features $f_{IA}$. The converted features and their corresponding label vectors are used to construct a concept-guided memory to represent the domain-specific visual knowledge. $W_T$ represents the semantic embeddings that are extracted from CLIP’s text encoder with handcrafted prompts of HOI labels, serving as the domain-agnostic semantic knowledge. All knowledge contained in the memory can be combined at inference time in a \textit{training-free} manner. During \textit{fine-tuning}, we unfreeze the properly initialized concept-guided memory and inject lightweight residual instance-aware adapters. The weights of blue blocks are frozen while the yellow blocks' are learnable. (Best viewed in color.)}
    \label{fig:Pipeline}
\end{figure*}

\section{Related Work}
\label{sec:related}

\subsection{HOI Detection}
HOI learning has been rapidly progressing in recent years with the development of large-scale datasets~\cite{gupta2015visual, chao2018learning, kuznetsova2020open, liao2020ppdm} and vision Transformers~\cite{vaswani2017attention, carion2020end}. HOI detection methods can be categorized into one- or two-stage approaches. 
One-stage HOI detectors~\cite{chen2021reformulating, kim2021hotr,tamura2021qpic,zou2021end,iftekhar2022look, qu2022distillation,wang2022learning,liao2022gen} usually formulate HOI detection task as a set prediction problem originating from DETR~\cite{carion2020end} and perform object detection and interaction prediction in parallel~\cite{chen2021reformulating,kim2021hotr, tamura2021qpic,zou2021end} or sequentially ~\cite{liao2022gen}. However, they tend to cost a lot of computing resources and converge slowly.
Two-stage methods~\cite{chao2018learning,qi2018learning,zhou2019relation,gao2020drg,zhang2021spatially,zhang2022efficient} usually utilize pre-trained detectors~\cite{ren2015faster, he2017mask,carion2020end} to detect human and object proposals and exhaustively enumerate all possible human-object pairs in the first stage. Then they design an independent module to predict the multi-label interactions of each human-object pair in the second stage. 
Despite their improved performance, both one- and two-stage methods suffer from sheer drop on rare classes, as shown in Figure~\ref{fig:rare_nonrare}, due to the long-tailed distribution of HOI training data. 

The most relevant work to ours is GEN-VLKT~\cite{liao2022gen}, which proposes a Visual-Linguistic Knowledge Transfer(VLKT) module to transfer knowledge from a visual-linguistic pre-trained model CLIP. Different from previous methods, our \modelname utilizes a concept-guided memory module to explicitly memorize balanced prototypical feature representations for each category, which not only helps quickly retrieve knowledge in the training-free mode, but also relieves forgetfulness for rare classes during fine-tuning.

\subsection{Adapting Pretrained Models}
Vision and language Pretraining is a fundamental research topic in artificial intelligence. It has made rapid progress~\cite{li2020oscar, radford2021learning, li2021align, kim2021vilt} in recent years thanks to the transformer mechanism~\cite{vaswani2017attention}. 
Instead of fine-tuning whole pre-trained models on downstream tasks which always costs a large of computational resources, many works study feature adapters~\cite{houlsby2019parameter,gao2021clip, zhang2021tip, sung2022vl, chen2022adaptformer} or prompt learning~\cite{zhou2022learning, jia2022vpt} to conduct fine-tuning. 

TipAdapter~\cite{zhang2021tip} propose to construct an adapter via a key-value cache model, where the idea of our concept-guided memory originates from. However, TipAdapter is designed for object classification and only needs to store object-level features. 
In contrast, our approach differs in two key aspects:
First, our approach is tailored for the more complex task of HOI understanding, and hence we store more fine-grained prototypical features including both instance-wise and pair-wise features.
Second, given the combination of the long-tailed human distribution and long-tailed object distribution, we balance the cached features of different HOI concepts. By doing so, we can ensure that each HOI concept is given appropriate weight and attention, even if some concepts are less frequent in the dataset.

Some works~\cite{gao2021clip, zhang2021tip} utilize adapter at the last layers for classification task, however, the valuable cues for HOI detection task may appear at the start from the early spatial-aware and fine-grained feature maps, which make any adjustment purely at the tail of the network less effective. 
Other works~\cite{sung2022vl, chen2022adaptformer} utilize adapter at the early layers of the pretrained visual encoders, but they do not focus on the pair-wise spatial relationship and thus are not suitable for the HOI detection task.
To address these we first propose to store features representing interaction concepts under training-free mode. Furthermore, we propose to not only unfreeze the global knowledge cache memory in fine-tuning, but also early injects prior knowledge to low-level feature maps to capture the local geometric spatial structures required for pair-wise relationship detection.

\section{Method}
\label{sec:method}

\subsection{Overview}
\label{subsec:overview}

The architecture of our \modelname is shown in Figure~\ref{fig:Pipeline}. Our approach is two-stage and consists of two main steps: 1) object detection and 2) interaction prediction. Given an image $\mathcal{I}$, we first use an off-the-shelf object detector $\mathcal{D}$, \eg DETR, and apply appropriate filtering strategies to extract all detections and exhaustively enumerate human-object pairs. Every human-object pair could be represented by a quintuple: $(b_{h}, s_{h}, b_{o}, s_{o}, c_{o})$. Here $b_{h}, b_{o} \in \mathbb{R}^{4}$ denote the detected bounding box of a human and object instance, respectively. $c_{o} \in \{1,...,O\}$ is the object category. $s_{h}$ and $s_{o}$ denote the confidence score of the detected human and object, respectively. The second stage includes a multi-branch concept-guided memory module (Figure~\ref{fig:Pipeline}(right)), to recognize the action category $a_{h,o} \in \{1,...,A\}$ and produce a confidence score $s_{h,o}^{a}$ for each human-object pair. 

\textit{In the training-free setting}, \modelname is well-suited for situations where there is limited data available. This makes it a valuable tool for addressing the challenges of data scarcity. We construct the multi-branch concept-guided memory, which consists of the instance-centric branch, the interaction-aware branch and the semantic branch, to store domain-specific visual knowledge and domain-agnostic semantic knowledge.
Given human-object pairs in a training set, we first extract the fine-grained features and interaction-aware features, for the former two branches. Fine-grained features are used to capture the detailed characteristics of HOIs, such as the pose or orientation state of the detected instances. On the other hand, interaction-aware features capture interaction-relevant environmental and contextual information that can affect the interaction, such as spatial layout and social context. Then for the semantic branch, we extract the semantic embeddings of the HOI categories and incorporate them into HOI cache memory as domain-agnostic knowledge. Semantic features enable the model to leverage linguistic commonsense in order to capture potential co-occurrences and affordance relations between objects and interactions.

\begin{figure}[t]
\begin{center}
   \includegraphics[width=0.8\linewidth]{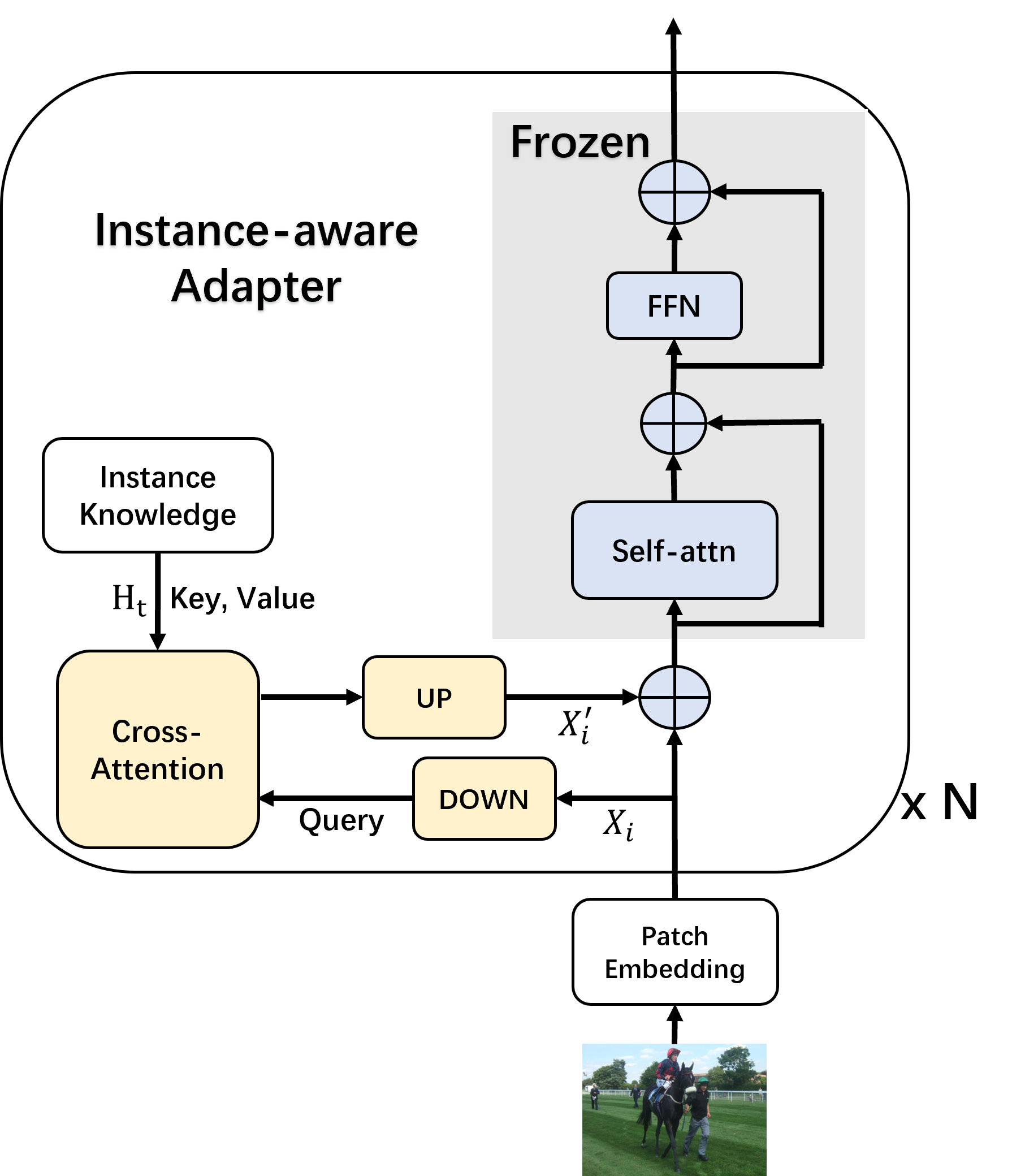}
\end{center}
    \caption{\textbf{Instance-aware Adapter.} Architecture of our modified CLIP visual encoder which is equipped with instance-aware adapter modules in each transformer block. 
    The instance-aware adapters aid in the early injection of instance-level prior knowledge into spatial-aware and fine-grained feature maps, thereby assisting the pair-wise relationship detection task.
    }
    \label{fig:Prior}
\end{figure}

\textit{In the fine-tuning setting}, to further improve the model's performance, we include an instance-aware adapter, as shown in Figure~\ref{fig:Prior}. This adapter injects prior knowledge into the visual encoder to better encode instance-level features through an effective fine-tuning. In addition, we unfreeze the cached keys of the concept-guided memory (yellow modules in Figure~\ref{fig:Pipeline}). The logits of these different types of memories are then linearly combined to output the final score $s_{h,o}^{a}$. Finally, the HOI score $\hat{s}_{h,o}^{a}$ for each human-object pair can be written as:

\begin{equation}
    \label{eq:supress}
    \hat{s}_{h,o}^{a} = (s_h \cdot s_o)^{\lambda} \cdot \sigma(s_{h,o}^{a})
\end{equation}
where $\lambda$ is a hyperparameter used at inference time to suppress overconfident objects\cite{zhang2021spatially} and $\sigma$ is a sigmoid function.

\subsection{Concept-guided Memory}
\label{subsec:memory}
Most conventional methods directly apply a multi-label classifier fitted from the dataset to recognize the HOIs for interaction understanding. However, due to the complicated human-centric scenes with various interactive objects, such paradigms suffer from a long-tailed distribution which is the combination of the long-tailed human distribution and the long-tailed object distribution. To alleviate the problem, we introduce a multi-branch concept-guided memory to explicitly store different types of  balanced concept features for all HOI classes. The concept-guided memory consists of three memory branches that are leveraged at the interaction prediction stage: 1) the instance-centric branch; 2) the interaction-aware branch; 3) the semantic branch. 
The three branches store interactive instances' appearance, interaction-relevant context, and linguistic common sense of each HOI class. These branches complement each other, enhancing the model's HOI understanding.
Among them, the instance-centric branch and the interaction-aware branch store the domain-specific visual knowledge while the semantic branch stores the domain-agnostic semantic knowledge.

\noindent \textbf{Instance-Centric Branch}
We use a key-value structure $(F_{IC}, L_{IC})$ for the Instance-Centric Branch as shown in Figure~\ref{fig:Pipeline}. We use $f_h, f_o$ to represent human features and object features, respectively. Given human-object pairs $(b_{h}, b_{o})$ in the training set, we use the CLIP to encode the cropped regions as $f_h, f_o$ for $b_h, b_o$, respectively. Then the concatenated features $f_{IC}$ = $Concat(f_h, f_o)$ are stored in $F_{IC}$ as a key and the corresponding labels are transformed as multi-hot vectors and stored in $L_{IC}$ as values.

\noindent \textbf{Interaction-Aware Branch}
Similar to the instance-centric branch, the interaction-aware branch is also modeled by a concept-feature dictionary $(F_{IA}, L_{IA})$. Given human-object pairs $(b_{h}, b_{o})$, we compute the union regions of the human-object pairs $b_{u}$ and extract the interaction-aware features $f_{IA}$ from the cropped union regions. Then $f_{IA}$ and its corresponding label vectors are stored in $(F_{IA}, L_{IA})$. The interaction-relevant information concentrates more on the background semantics and the spatial configuration of human-object pairs. This design choice further help to make more informed interaction predictions.

\noindent \textbf{Semantic Branch}
While the prior two branches are rich in domain-specific knowledge, the semantic branch focuses on domain-agnostic knowledge, leveraging linguistic common sense to boost interaction prediction.
To construct $W_T$ for the semantic branch, we first use handcrafted prompts (i.e., A photo of a person is $<$ACTION$>$ an object) to generate the raw text description of interactions. Then we pass such query through the CLIP text encoder and obtain the weights for the semantic classifier. Note that the semantic branch utilizes the alignment of visual and textual feature generated by CLIP, and thus can easily extend to a novel HOI by adding its semantic feature to the classifier.

Given the above branches of concept-guided memory, the interaction scores could be estimated as follows:
\begin{equation}
    \label{eq:similarity}
    \begin{aligned}
        s_{vis} &= \gamma_{IC} \cdot (f_{IC} F_{IC}^T) L_{IC} + \gamma_{IA} \cdot (f_{IA} F_{IA}^T) L_{IA} \\
        s_{h,o}^{a} &= s_{vis} + \gamma_{T} \cdot f_{U} W_{T}^T \\
    \end{aligned}
\end{equation}
where $\gamma_{IC}$, $\gamma_{IA}$ and $\gamma_{T}$ controls the balanced weights of the three branches, and $f_U$ represents the feature of union regions of human-object pairs, identical to $f_{IA}$ during implementation.

\subsection{Instance-aware Adapter}
\label{subsec:adapter}

\modelname could utilize the knowledge from CLIP and a few-shot training set to achieve competitive performance for HOI detection tasks in the training-free setting. However, the performance improvement upper-bounds as more and more samples are cached as shown in Figure~\ref{fig:ablation_memory_shot}. 
The information stored in memory features is a holistic representation that lacks detailed spatial information, which is critical for HOI understanding.
As a result, we design an instance adapter that could be inserted into each block of the CLIP visual encoder as shown in Figure~\ref{fig:Prior}. By incorporating the prior knowledge of the feature maps rich in spatial configurations, we equip CLIP with instance-aware knowledge, making it better understand actions to boost the performance of interaction prediction.

\noindent\textbf{Prior Knowledge} The prior knowledge consists of three components. (1) Spatial configuration, which captures the geometric information of objects and enable discrimination between fine-grained categories. (2) Semantic information of extracted objects: we use CLIP language encoder to obtain visual embeddings for given detected objects, which enables us to leverage the language priors to capture which objects can be interacted with. (3) A confidence score that reflects the quality/uncertainty of the candidate instance. 
The prior knowledge is unique to each image and contains information about all the detected instances in it.
The process of the instance-aware adapter, which fuses the CLIP's visual features with these components, will be introduced below.

\noindent \textbf{Instance-aware Adapter}
The instance-aware adapters are sub-networks with a small number of parameters and replaced at the beginning of every block in the transformer encoder. With the adapters, \modelname could learn prior knowledge from extracted objects which provides better instance representation for downstream tasks with fine-tuning. We use cross-attention module to inject the prior knowledge to CLIP's visual encoder as shown in Figure~\ref{fig:Prior}.

To be more specific, we denote $X_{i} \in \mathbb{R}^{H'W' \times d}$ as the input feature map of i-th block of the visual encoder, where $H'W'$ is the shape of the feature map. To reduce the computational cost, we first reduce the dimension of features through down-projection. The weight matrices of down-projection and up-projection blocks are denoted as $W_{down} \in \mathbb{R}^{d \times d^{\prime}}$ and $W_{up} \in \mathbb{R}^{d^{\prime} \times d}$, where $d$ and $d^{\prime}$ satisfy $d^{\prime} << d$. Additionally, we append a multi-head cross attention ($MHCA$) module after the first projection layer to incorporate the outputs of the down-projection layer and the prior knowledge from DETR:
\begin{equation}
  \left.\begin{aligned}
        H_t &= MLP_{p}(P_t) \\
        X_{i}' &= MHCA((X_i \cdot W_{down}), H_t, H_t) \cdot W_{up}\\
  \end{aligned}
  \right.
\end{equation} 
where $P_t = \{p_{t}\}_{t=1}^{N_t}$ represents the prior knowledge of extracted instances, $p_t = \{b_t, c_t, e_t \} $ consists of box coordinates $b_t$, the confident score $c_t$, the object text embedding $e_t$ extracted from CLIP text encoder and $N_t$ denotes the number of extracted instances in a given image. $H_t = \{h_{t}|h_{t} \in \mathbb{R}^{d^{\prime}}\}_{t=1}^{N_t}$ represents hidden features obtained after the prior features pass through down-projection MLPs.


\begin{table*}
\begin{center}
\begin{tabular}{|l|l|c|ccc|}
\hline
Method & Backbone & TP & Full  &  Rare   &  Non-rare \\
\hline\hline
\multicolumn{6}{l}{\textbf{One-stage Methods}} \\
    GPNN~\cite{qi2018learning} & ResNet-101 & & 13.11& 9.41 & 14.23 \\
    UnionDet~\cite{kim2020uniondet} & ResNet-50-FPN & - & 17.58 & 11.72 & 19.33\\
     IP-Net~\cite{wang2020learning} & Hourglass-104 & -& 19.56 & 12.79 & 21.58 \\
     PPDM~\cite{liao2020ppdm} & Hourglass-104 &- & 21.94  & 13.97   &  24.32 \\
     HOI-Trans~\cite{zou2021end} &  ResNet-50 &- & 23.46  & 16.91 & 25.41  \\
     HOTR~\cite{kim2021hotr}   &  ResNet-50 & 9.90 & 25.10 & 17.34 &  27.42  \\
     AS-Net~\cite{chen2021reformulating}  & ResNet-50 & -&   28.87  & 24.25 & 30.25 \\
     QAHOI~\cite{chen2021qahoi} & Swin-Base &- & 29.47 & 22.24 & 31.63\\
     QPIC~\cite{tamura2021qpic} &  ResNet-101 & 41.46 & 29.90 &23.92 &31.69 \\
    Iwin~\cite{tu2022iwin} & ResNet-50-FPN & -& 32.03 & 27.62 & 34.14 \\
    CDN~\cite{zhang2021mining} & ResNet-101 & -& 32.07 & 27.19 & 33.53 \\
    GEN-VLKT~\cite{liao2022gen}  & ResNet-50+ViT-B & 42.05 & 33.75 & 29.25 & 35.10 \\
\hline\hline
    \multicolumn{6}{l}{\textbf{Two-stage Methods}} \\
     InteractNet~\cite{gkioxari2018detecting} & ResNet-50-FPN & -&  9.94 & 7.16 & 10.77 \\
     iCAN~\cite{gao2018ican}  & ResNet-50 & - & 14.84   &10.45  &16.15  \\
     TIN~\cite{li2019transferable} & ResNet-50 & - & 17.03 & 13.42 &18.11 \\
     PMFNet*~\cite{wan2019pose}  &  ResNet-50-FPN &- & 17.46  &15.65 &18.00  \\ 
     DRG~\cite{gao2020drg}   & ResNet-50-FPN &- & 19.26 & 17.74 &19.71  \\
     VCL~\cite{hou2020visual}    &  ResNet-50 & - & 19.43  & 16.55  &20.29   \\
     FCMNet*~\cite{liu2020amplifying}  & ResNet-50 & -& 20.41 &17.34  &21.56  \\
     ACP~\cite{kim2020detecting} &   ResNet-152 & -& 20.59  &15.92  &21.98   \\
     IDN*~\cite{li2020hoi}  &  ResNet-50 &- &23.36  &22.47   &23.63    \\
     STIP~\cite{zhang2022exploring} & ResNet-50 & - & 30.56 & 28.15 & 31.28 \\
     UPT (TF)~\cite{zhang2022efficient} $\dagger$ & ResNet-50 & 0 & 13.64 & 9.66 & 14.85 \\
     UPT~\cite{zhang2022efficient} & ResNet-50 & 13.24 & 31.66 & 25.94 & 33.36\\ 
     \modelname (TF) & ResNet-50+ViT-B & 0 & 25.19 & 27.24 & 24.58 \\
     \modelname (FT) & ResNet-50+ViT-B & 3.12 & 33.80 & 31.72 & 34.42 \\
     \modelname (FT) & ResNet-50+ViT-L & 6.62 & \textbf{38.40} & \textbf{37.52} & \textbf{38.66} \\

\hline
\end{tabular}
\end{center}
\caption{\textbf{State-of-the-art Comparison on HICO-DET.} The table compares the HOI detection performance (mAP×100) on the HICO-DET test set. TP: Number of Trainable Params(M), TF: Training-Free, FT: Fine-Tuning. "*" indicates the method uses additional features from pose estimation of body-parts. $\dagger$ indicates we
apply our generic memory design to a representative approach. 
Our method equipped with fine-tuned adapter achieves a new state-of-the-art.
}
\label{tab:hicodet}
\end{table*}


\subsection{Training and Inference}
\label{subsec:training_and_inference}

\noindent \textbf{Training-Free Setting}
In the training-free setting, given an image at inference time, for every detected human-object pair, we extract its features $f_h, f_o, f_u$ as presented in Section~\ref{subsec:memory}. Then the interaction scores $\hat{s}_{h,o}^{a}$ could be calculated as shown in Equation~\ref{eq:supress} and ~\ref{eq:similarity}.

\noindent \textbf{Fine-Tuning Setting}
During the fine-tuning phase, the blue components shown in Figure~\ref{fig:Pipeline} are frozen and kept the same as those in the training-free phase while the weights of yellow components are trainable. 
The process of initializing the cache memory is identical to that used in the training-free mode. 
We use the original image as the input of visual encoder and adopt ROI-Align\cite{he2017mask} to extract $f_{IC}$, $f_{IA}$ defined in Section~\ref{subsec:memory}. The whole model is trained on focal loss~\cite{lin2017focal} $\mathcal{L}$. 
We denote $\theta$ as the learnable parameters of our model.
The optimization procedure could be written as follows:
\begin{equation}
    \theta^{*} = \mathop{\arg\min}\limits_{\theta} \mathcal{L}( g( f(\mathcal{D}(\mathcal{I}), \mathcal{I}), M) , s_{GT}) \\
    \label{eq:optimization}
\end{equation}
where $\mathcal{I}$, $\mathcal{D}$ represent the input image and pre-trained detector as defined in Section~\ref{subsec:overview}, $f$ is the feature extraction procedure presented in Section~\ref{subsec:memory}, $M$ represents the concept-guided memory, $g$ is a similarity function defined in Equation~\ref{eq:similarity}, and $s_{GT}$ represents the ground-truth label.


\section{Experiments}
\label{sec:experiments}

\subsection{Experimental Settings}
\label{subsec:exp_setting}
\noindent \textbf{Datasets:} We conducted extensive experiments on both the HICO-DET~\cite{chao2018learning} and V-COCO~\cite{gupta2015visual} datasets. HICO-DET consists of 47,776 images (38,118 training images and 9,658 test images). It has 600 HOI categories constructed of 117 action classes and 80 object classes. 
In addition, we simulate a zero-shot detection setting by holding out 120 rare interactions following previous settings~\cite{hou2020visual, hou2021affordance, wang2022learning}.
V-COCO is subset of COCO and has 10,396 images (5,400 trainval images and 4,946 test images). It has 24 different types of actions and 80 types of objects. 

\noindent \textbf{Evaluation Metric:} Following the standard evaluation, we use mean average precision (mAP) to examine the model performance.
For HICO-DET, we report the mAP over three different category sets: all 600 HOI categories (Full), 138 HOI categories with less than 10 training instances (Rare), and the remaining 462 HOI categories (Non-rare). For V-COCO, we report the average precision (AP) under two scenarios, $AP_{role}^{S1}$ and $AP_{role}^{S2}$, which represent different scoring ways for object occlusion cases.

\subsection{Implementation Details}
We fine-tune the detector DETR prior to training and then freeze its weights. Specifically, for HICO-DET, we fine-tune DETR on HICO-DET with its weights initialized from the publicly available model pre-trained on MS COCO~\cite{lin2014microsoft}. For V-COCO, we pre-train DETR from scratch on MS COCO, excluding those images in the test set of V-COCO. 
We employ two ViT variants as our backbone architectures: ViT-B/16 and ViT-L/14, where "B" and "L" refer to base and large, respectively. 
The input resolution for ViT-B and ViT-L is 224 pixels and 336 pixels, respectively.
$\gamma_{IC}$, $\gamma_{IA}$ and $\gamma_{T}$ are set to be 0.5, 0.5 and 1.0, respectively. 
$\lambda$ is set to 1 during training and 2.8 during inference~\cite{zhang2021spatially, zhang2022efficient}. We use AdamW~\cite{loshchilov2017decoupled} as the optimizer with an initial learning rate of 1e-3 and train \modelname for 15 epochs. The model is trained on a single NVIDIA A100 device with an efficient batch size of 8.

\begin{table}
\begin{center}
\begin{tabular}{|l|cc|}
\hline
    Method &  $AP_{role}^{S1}$   &  $AP_{role}^{S2}$  \\
\hline\hline
\multicolumn{3}{l}{\textbf{One-stage Methods}} \\
    UnionDet~\cite{kim2020uniondet}  & 47.5 & 56.2  \\
    IP-Net~\cite{wang2020learning} & 51.0 & -  \\
    HOI-Trans~\cite{zou2021end}  & 52.9 & - \\
    GG-Net~\cite{zhong2021glance}  & 54.7 & - \\
    HOTR~\cite{kim2021hotr} & 55.2 & 64.4 \\
    AS-Net~\cite{chen2021reformulating} & 53.90 & - \\
    QPIC~\cite{tamura2021qpic}  & 58.8 &  61.0 \\
    Iwin~\cite{tu2022iwin} & 60.47 & - \\
    CDN\ding{61}~\cite{zhang2021mining} & 61.68 & 63.77 \\
    GEN-VLKT\ding{61}~\cite{liao2022gen} & \textbf{62.41} & 64.46 \\ 
    
\hline\hline
    \multicolumn{3}{l}{\textbf{Two-stage Methods}} \\
    InteractNet~\cite{gkioxari2018detecting}&  40.0 & - \\
     iCAN~\cite{gao2018ican}&  45.3 &  52.4   \\
     PMFNet*~\cite{wan2019pose}&  52.0 & -  \\ 
     VCL~\cite{hou2020visual}& 48.3 &  -  \\
     DRG~\cite{gao2020drg}& 51.0  & -  \\
     FCMNet*~\cite{liu2020amplifying}  & 53.1  & - \\
     ACP*~\cite{kim2020detecting} & 53.2  & - \\
     IDN*\ding{61}~\cite{li2020hoi} & 53.3 &  60.3 \\
     UPT~\cite{zhang2022efficient} & 59.0 & \textbf{64.5} \\ 
     \modelname (TF, ViT-B) & 39.09 & 43.93 \\
     \modelname (FT, ViT-B) & 56.12 & 61.45 \\
     \modelname (FT, ViT-L) & 58.57 & 63.97 \\
\hline
\end{tabular}
\end{center}
\caption{\textbf{State-of-the-art comparison on V-COCO.} We report HOI detection performance (mAP×100) on the V-COCO test set. "*" indicates the method uses additional features from pose estimation. \ding{61} indicates the method uses bounding box annotations of test set to pre-train the object detector. Note that our method achieves competitive performance \textit{wrt} the state-of-the-art. It is also worth highlighting that some of these methods require extensive training and might be over-specialized for this target dataset.}
\label{tab:vcoco}
\end{table}

\begin{table}
\begin{center}
 \begin{tabular}{|c|c|ccc|}
    \hline
    Method & Type & Unseen & Seen & Full \\
    \hline \hline
      VCL~\cite{hou2020visual} & RF & 10.06 & 24.28 & 21.43 \\
      ATL~\cite{hou2021affordance} & RF & 9.18  & 24.67 & 21.57 \\
      FCL~\cite{hou2021detecting} & RF & 13.16 & 24.23 & 22.01 \\
      THID~\cite{radford2021learning} & RF & 15.53 & 24.32 & 22.96 \\
      GEN-VLKT~\cite{liao2022gen} & RF & 21.36 & 32.91 & 30.56 \\
      Ours (TF, ViT-B) & RF & 26.83 & 24.54 & 25.00 \\
      Ours (FT, ViT-B) & RF & \textbf{27.63} & \textbf{34.35} & \textbf{33.01} \\ 
    \hline \hline
      VCL~\cite{hou2020visual} & NF & 16.22 & 18.52 & 18.06 \\
      ATL~\cite{hou2021affordance} & NF & 18.25 & 18.78 & 18.67 \\
      FCL~\cite{hou2021detecting} & NF & 18.66 & 19.55 & 19.37 \\
      GEN-VLKT~\cite{liao2022gen} & NF & 25.05 & 23.38 & 23.71 \\
      \modelname (TF, ViT-B) & NF & 30.11 & 23.16 & 24.55 \\
      \modelname (FT, ViT-B) & NF & \textbf{32.41} & \textbf{31.13} & \textbf{31.39} \\
    \hline
\end{tabular}
\end{center}
\caption{\textbf{Zero-shot comparison on HICO-DET.} This table compares the performance of our model with state-of-the-art methods on the Zero-shot setting of HICO-DET. RF: Rare First. NF: Non-rare First. TF: Training-Free. FT: Fine-Tuning. 
} 
\label{tab:ZS}
\end{table}

\begin{table}
\begin{center}
\begin{tabular}{|c|c|c|ccc|}
    \hline
    IC & IA & S & Full & Rare & Non-rare \\
    \hline\hline
    \ding{51} &  &  & 22.36 & 21.38 & 22.65 \\
     & \ding{51} &  & 21.74 & 20.40 & 22.15 \\
     &  & \ding{51} & 23.36 & \textbf{27.81} & 22.03 \\
     & \ding{51} & \ding{51} & 25.02 & 27.28 & 24.35 \\
    \ding{51} & \ding{51} & \ding{51} & \textbf{25.19} & 27.24 & \textbf{24.58} \\
    \hline
\end{tabular}
\end{center}
    \caption{\textbf{Concept-guided Memory ablation (Training-Free).} This table studies the effect of each branch on the training-free setting. IC: Instance-Centric branch, IA: Interaction-Aware branch, S: Semantic branch. Results are on HICO-DET.} 
\label{tab:ablation_branch_TF}
\end{table}

\begin{table}
\begin{center}
\begin{tabular}{|c|c|c|c|ccc|}
    \hline
        Adapter & IC & IA & S & Full & Rare & Non-rare \\
        \hline\hline
         & \ding{51} & \ding{51} & \ding{51} & 27.63 & 25.40 & 28.30 \\
        \ding{51} & \ding{51} & & & 32.95 & 32.32 & 33.13 \\
        \ding{51} &  & \ding{51} & & 31.26 & 30.77 & 31.40 \\
        \ding{51} &  &  & \ding{51} & 30.48 & 29.12 & 30.89 \\
        \ding{51} & \ding{51} & \ding{51} & \ding{51} & \textbf{33.80} & \textbf{31.72} & \textbf{34.42} \\
    \hline
    \end{tabular}
\end{center}
\caption{\textbf{Concept-guided Memory ablation (Fine-Tuning).} This table studies the effect of each branch on the fine-tuning setting. IC: Instance-Centric branch, IA: Interaction-Aware branch, S: Semantic branch, Adapter: Instance-aware Adapter. Results are on HICO-DET.} 
\label{tab:ablation_branch_F}
\end{table}

\begin{figure}[h]
\begin{center}
   \includegraphics[width=0.8\linewidth]{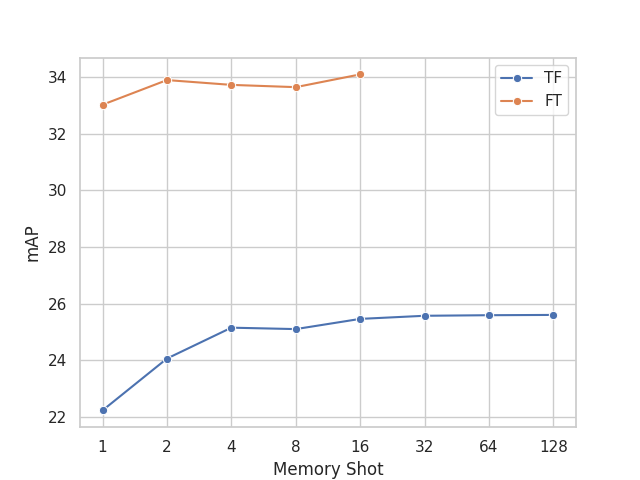}
\end{center}
    \caption{\textbf{Memory size ablation.} Memory shot indicates the number of samples in the memory per HOI.  TF: Training-Free, FT: Fine-Tuning.  We observe two behaviors: (i) FT performs well independently as to the memory size, and (ii) our TF benefits from larger memory sizes but does pretty well with as low as 16-32 samples.} 
\label{fig:ablation_memory_shot}
\end{figure}

\subsection{HOI State-of-the-art Comparison}
We compare the performance of our model with existing methods on HICO-DET and V-COCO datasets. 
For HICO-DET, we evaluate our model on both the default setting and the zero-shot setting as described in Section~\ref{subsec:exp_setting}. As shown in Figure~\ref{tab:hicodet}, our model under training-free setting outperforms many models on the HICO-DET dataset, \eg HOI-Trans, IDN, which require training. 
Note that we are the first to propose a training-free HOID approach. Existing one-stage/two-stage methods require training the decoder/classifier respectively. Thus none of them can directly provide results for comparison without training.
To further verify our method has advantages against others, we apply our generic memory design to a representative approach~(UPT). Empirically, as shown in Table~\ref{tab:hicodet}, we observe that UPT achieves a mAP of 13.64\%, while our method reaches 25.19\% on HICO-DET full split, demonstrating the effectiveness of our approach. Notably, transformer-based one-stage HOI detectors require training for instance grouping and pair-wise box prediction, which impedes the acquisition of reasonable training-free results.

Furthermore, our model(ViT-B) after fine-tuning outperforms all other methods, which shows our adapter's strong capability of modeling pair-wise human-object relationships. Our model(ViT-L) can further leverage high-resolution input images due to its computational efficiency, which also facilitates feature extraction for small objects and boosts its performance. To be specific, our model achieves 4.65\% and 6.74\% mAP gains compared to the state-of-the-art of one- and two-stage methods, respectively. It's worth noting that existing models tend to forget rare HOI classes while ours achieve a significant improvement especially on the Rare HOI categories, which confirms the effectiveness of our concept-guided memory for memorizing rare HOI concepts. 

We choose ViT as our backbone to leverage the aligned image-text feature space provided by CLIP. In Table~\ref{tab:hicodet}, we provide a fair comparison with GEN-VLKT, which uses the same pre-trained weights as our model. Our approach achieves better performance on rare classes, demonstrating the efficiency and effectiveness of our method. It is noteworthy that our model requires less training time(0.1$\times$), as illustrated in Figure~\ref{fig:scatter}.
Besides, the "Backbone" column in Table~\ref{tab:hicodet} stands for which part of the model is pre-trained. Most methods with the backbone of ResNet still have transformer encoder layers followed, and thus need to optimize a large amount of parameters during training. Though a stronger backbone (ViT-B) with a frozen detector, \modelname has a small number of trainable parameters.
As the main architecture of our backbone is frozen, it can easily be upgraded to more powerful one without struggling and have better scalability. The last row in Table~\ref{tab:hicodet} shows that our model's performance can boost a lot through an easy switch, where the overall trainable parameters (tailored for this dataset) are still relatively small.

The results presented in Table~\ref{tab:ZS} demonstrate the effectiveness of our proposed method for zero-shot HOI detection on HICO-DET. Our model achieves a relative gain of 29.35\% and 29.38\% mAP on the two zero-shot settings, respectively, compared to the best performing approach, GEN-VLKT. This improvement shows the great generalizability of our model for detecting HOIs belonging to unseen combinations and the ability to disentangle the concept of action and object.
It is also worth noting that under the Non-rare First setting, specifically the systematic-hard split which contains fewer training instances and is thereby more challenging, our model shows great advantages over the seen categories compared to GEN-VLKT. This demonstrates the effectiveness of domain-specific visual knowledge in our memory.
Overall, the performance of our model has significantly improved, demonstrating the effectiveness of the aligned image-text feature space.

For V-COCO, as shown in Table~\ref{tab:vcoco}, our model achieves competitive results compared with all previous methods which also freeze the weights of detector during training. The improvement is not that significant compared to HICO-DET, which might be caused by the insufficient training samples in V-COCO.

\subsection{Ablation Study}
In this subsection, we conduct a series of experiments to study the effectiveness of our proposed modules. All experiments are conducted on the HICO-DET dataset. 

\noindent \textbf{Memory size.}
Here we study the impact of the size of our cache model. As shown in Figure~\ref{fig:ablation_memory_shot}, in the training-free setting, as the memory shot increases from 1 to 16, the performance consistently improves. However, when we further expand the cache model size, the performance achieves an upper-bound around 26\%. For the fine-tuning setting, the performance is not sensitive to the memory size as we can see from the orange line in Figure~\ref{fig:ablation_memory_shot}, which shows the model's capability of efficiently utilizing the memory.   

\noindent \textbf{Network architecture.}
This ablation studies the effectiveness of different modules of our network. In the training-free setting, as shown in Table~\ref{tab:ablation_branch_TF}, by keeping one branch at a time, we could observe that the semantic branch contributes the most to the performance gains. We suppose the reason is that the domain-agnostic linguistic knowledge is well-aligned with the region features thanks to the zero-shot capability of large visual-language model. The Instance-centric branch and the Interaction-Aware branch further boost the performance, which demonstrates the complementarity of domain-specific visual knowledge and domain-specific linguistic language. In the lightweight tuning setting, we show the effectiveness of different components in Table~\ref{tab:ablation_branch_F}. Compared to the performance in training-free setting, simply unfreezing the keys of memory boosts the performance from 25.19\% to 27.63\%. By further utilizing our designed instance-aware adapter for better pair-wise relationship modeling, we achieve the best performance of mAP 33.80\%.

\begin{table}
\begin{center}
\footnotesize
\begin{tabular}{|cc|ccc|}
    \hline
    DS & DA & Full & Rare & Non-rare \\
    \hline \hline
    \ding{51} &  & 32.92 & 31.36 & 33.39 \\
      & \ding{51} & 30.48 & 29.12 & 30.89 \\
    \ding{51} & \ding{51} & \textbf{33.80} & \textbf{31.72} & \textbf{34.42} \\
    \hline
\end{tabular}
\end{center}
    \caption{\textbf{Ablation on different types of knowledge.} This table studies the effect of domain-agnostic and domain-specific knowledge on the fine-tuning setting. DS: Domain-Specific. DA: Domain-Agnostic. Results are on HICO-DET.} 
\label{tab:DS-DA}
\end{table}

\noindent \textbf{Different types of knowledge.}
As shown in Table~\ref{tab:DS-DA}, we also group the branches by domain-agnostic and domain-specific knowledge to see the effect of the different types of knowledge on the performance of the method in the Fine-Tuning mode.
Experimental results demonstrate that domain-specific knowledge customized for the specific dataset outperforms domain-agnostic knowledge by an obvious margin of 2.24\% and 2.5\% mAP on rare and non-rare splits, respectively.
By further taking advantage of their complementary knowledge, the combination achieves the best overall performance on all the split of HICO-DET.


\section{Conclusion}
\label{sec:conclusion}

We proposed \modelname, an efficient adaptive model for Human-Object Interaction Detection. Our method showed its effectiveness in both training-free and fine-tuning operation modes. In the training-free setting, our \modelname achieves a competitive performance by constructing a concept-guided cache memory that leverages domain-specific visual knowledge and domain-agnostic semantic knowledge. Furthermore, in the fine-tuning setting, with the designed instance-aware adapter, our model achieves competitive results with sota on zero-shot and default setting. 

\section{Acknowledgements}
This work was supported by National Natural Science Foundation of China (61925201,62132001), Zhejiang Lab (NO.2022NB0AB05), Adobe and CAAI-Huawei MindSpore Open Fund. We thank MindSpore\footnote{https://www.mindspore.cn/}
for the partial support of this work, which is a new
deep learning computing framework.



{\small
\bibliographystyle{ieee_fullname}
\bibliography{egbib}
}

\end{document}